\documentclass[extendedabs]{recpad2k}
\usepackage[named]{algo}
\usepackage{graphics}
\usepackage{graphicx}
\usepackage{stfloats}
\usepackage{amssymb}
\usepackage{amsmath}
\usepackage{amsfonts}
\usepackage{subfigure}
\usepackage{float}
\usepackage{pifont}
\usepackage{setspace}

\usepackage{url}
\title{Exploiting Group Sparsity in 2D 1-Bit Compressive Sensing}

\addauthor{Xiangrong Zeng}{Xiangrong.Zeng@lx.it.pt}{1}
\addauthor{M\'{a}rio A. T. Figueiredo}{mario.figueiredo@lx.it.pt}{1}

\addinstitution{
 Instituto de Telecomunicações\\
 Instituto Superior Técnico\\
 Lisbon, Portugal
}

\runninghead{Student, Prof, Collaborator}{RECPAD Author Guidelines}

\begin{document}

\maketitle

\begin{abstract}
We propose a new approach, {\it two-dimensional binary fused compressive sensing} (2DBFCS) to recover
 2D sparse piece-wise signals from 1-bit measurements, exploiting group sparsity in 2D 1-bit compressive sensing. 
 The proposed method is a modified 2D version of the previous {\it binary iterative hard 
thresholding} (2DBIHT) algorithm, where, the objective function consists of a 2D one-sided $\ell_1$ 
(or $\ell_2$) function and an indicator function of $K$-sparsity and an indicator function of total variation 
(TV) or modified TV (MTV) constraint (the MTV favors both sparsity and piece/wise smoothness while the TV 
promotes the whole smoothness). 
The subgradient of 2D one-sided $\ell_1$ (or $\ell_2$) barrier and the projection onto the $K$-sparsity 
and TV or MTV constraint set are easy to compute, such that  the 
{\it forward-backward splitting} can be applied in 2DBFCS efficiently.
Experiments on the recovery of 2D sparse piece-wise smooth signals show that the proposed 2DBFCS with the TV or MTV
is able to take advantage of the piece-wise smoothness of the original signal, achieving more accurate recovery than
2DBIHT. Especially, the 2DBFCS with the MTV and the  $\ell_2$ barrier performs best amongst the algorithms. 
\end{abstract}

\section{Introduction}
We focus on recovering a sparse piece-wise smooth
two-dimensional (2D) signal (an image) ${\bf X}$ from 1-bit measurements,
\begin{equation}\label{1bitcsnoisy}
 {\bf Y}=\mbox{sign}\left({\bf A}{\bf X} + {\bf W}\right),
\end{equation}
where ${\bf Y}\in\{-1,1\}^{M \times L}$ is the measurement matrix, $\mbox{sign}$ is the element-wise 
sign function that returns $+1$ for positive arguments and $-1$ otherwise, ${\bf A}\in\mathbb{R}^{M\times N}$ is 
the known sensing matrix, ${\bf X} \in {\mathbb R}^{N \times L}$ is the original 2D signal, and ${\bf W} $ is additive noise.
 Unlike in conventional {\it compressive sensing} (CS), 1-bit measurements loose any information about the magnitude of
 the original signal ${\bf X}$. The goal is then to recover ${\bf X}$, but only up to an unknown and unrecoverable magnitude 
\cite{boufounos20081}, \cite{jacques2011robust}.

Our innovations, with respect to 1-bit CS as proposed in \cite{boufounos20081}, are twofold:
a) we address  the 2D case; b) more importantly, we introduce a new regularizer favoring both sparsity and
piece-wise smoothness, which can be seen as a modified 2D version of {\it fused lasso} \cite{tibshirani2004sparsity}. 
This new regularizer is the indicator of a union of convex subsets (total-variation balls) of the canonical subspaces,  
simultaneously enforcing sparsity and smoothness within each connected subset of
non-zero elements. The rationale is that, when imposing smoothness and sparseness, smoothness should not interfere with
sparsity, {\it i.e.}, it should not be imposed across the transitions from zero to non-zero elements.
The proposed regularizer promotes sparseness and smoothness
and (although it is non-convex) has a computationally feasible projection,
based on which we propose a modified version of the {\it binary iterative hard
thresholding} (BIHT)  algorithm \cite{jacques2011robust}.

\section{2D Binary Iterative Hard Thresholding (2DBIHT)}

To recover ${\bf X}$ from ${\bf Y}$, we first consider a 2D version of the
 criterion proposed by Jacques {\it et al} \cite{jacques2011robust}
\begin{equation}
\label{2DBIHT_model}
\begin{split}
& \min_{\bf X}  f({\bf Y}\odot {\bf A}{\bf X}) + \iota_{\Sigma_K}({\bf X})\\
& \mbox{subject\; to }\; \left\|{\bf X} \right\|_2 = 1,
\end{split}
\end{equation}
where:  the operation ``$\odot$" denotes element-wise (Hadamard) product;
$\iota_C\left({\bf X}\right)$ denotes the indicator function of set $C$,
\begin{equation}
\iota_C\left({\bf X}\right) = \left\{
\begin{array} {ll}
0,	&  {\bf X} \in C \\
+\infty, & {\bf X} \not\in C;
\end{array}\right.
\label{indicator}
\end{equation}
${\Sigma_K} = \left\{{\bf X}\in \mathbb{R}^{N \times L} : \left\|{\bf X}\right\|_0 \leq K \right\}$ (with
$\|{\bf V}\|_0$ denoting the number of non-zero components in ${\bf V}$) 
is the set of $K$-sparse $N \times L$ images; $\|{\bf X}\|_2 =  \bigl(\sum_{ij} X_{(i,j)}^2\bigr)^{\frac{1}{2}} $ is the  Euclidean norm,
and $f$ is one of the penalty functions defined next.
To penalize linearly the violations of the sign consistency between the observations and the estimate
\cite{jacques2011robust}, the barrier function is chosen as $f({\bf Z}) = 2\left\|{\bf \left[Z\right]}_{-}\right\|_1$,
where ${\bf \left[Z\right]}_{-} = \min \left({\bf Z}, 0\right)$ (with the minimum
applied entry-wise and the factor 2 included for later convenience) and $\|{\bf V}\|_1 = \sum_{ij} |V_{(i,j)}|$ is the
$\ell_1$ norm of ${\bf V}$.  A quadratic barrier for sign violations (see \cite{boufounos20081})
is achieved by using  $f({\bf Z})=\frac{1}{2} \left\|{\bf \left[Z\right]}_{-}\right\|_2^2 $, 
where the factor $1/2$ is also included for convenience.
The iterative hard thresholding (IHT) \cite{blumensath2009iterative} algorithm applied to \eqref{2DBIHT_model}
(ignoring the norm constraint during the iterations) leads to the 2DBIHT algorithm, which is a 2D version of the {\it binary iterative hard thresholding} (BIHT) \cite{jacques2011robust}:  
\begin{algorithm}{2DBIHT}{
\label{alg:2DBIHT}}
set $k =0, \tau >0, {\bf X}_0$ and $K$ \\
\qrepeat\\
     ${\bf V}_{k+1} = {\bf X}_{k} - \tau {\partial} f\left({\bf Y}\odot \left({\bf A}{\bf X}_k\right)\right)$\\
		 ${\bf X}_{k+1} =  {\mathcal P}_{\Sigma_K} \left( {\bf V}_{k+1}\right)$\\
		 $ k \leftarrow k+ 1$
\quntil some stopping criterion is satisfied.\\
\qreturn ${\bf X}_k/\left\|{\bf X}_k\right\|_2$
\end{algorithm}

In this algorithm, ${\partial} f$
denotes the subgradient of the objective (see \cite{jacques2011robust}, for details),
which is given by
\begin{equation} \label{subgradient}
{\partial} f\left({\bf Y}\odot \left({\bf A}{\bf X} \right)\right) = \left\{
\begin{array} {ll}
{\bf A}^T\left(\mbox{sign}\left({\bf A}{\bf X} \right) - {\bf Y}\right),	& \ell_1 \;\mbox{barrier,} \\
{\bf A}^T \left( {\bf Y}\odot\left[{\bf Y}\odot \left({\bf A}{\bf X} \right)\right]_- \right), & \ell_2 \;\mbox{barrier,}
\end{array}\right.
\end{equation}
Step 3 corresponds to a sub-gradient descent step (with step-size $\tau$), while
Step 4 performs the projection onto the non-convex set ${\Sigma_K}$, which 
corresponds to computing the best $K$-term approximation of ${\bf V}$, that is, 
keeping $K$ largest components in magnitude and setting the others to zero. Finally, the 
returned solution is projected onto the unit sphere to satisfy the constraint $\left\|{\bf X} \right\|_2 = 1$ in (\ref{2DBIHT_model}).
The versions of BIHT with $\ell_1$ and $\ell_2$ objectives are referred to as 2DBIHT-$\ell_1$ and 2DBIHT-$\ell_2$, respectively.

\section{2D Fused Binary Compressive Sensing (2DFBCS)}
The proposed formulation essentially adds a new constraint of
low (modified) total variation to the criterion of 2DBIHT \eqref{2DBIHT_model}, which encourages
4-neighbor elements to be similar, justifing the term ``fused". 

\subsection{2DFBCS with Total Variation}
We first propose the following model:
\begin{equation}\label{2DBFCS_model1}
\begin{split}
& \min_{\bf X}  f\left({\bf Y}\odot {\bf A}{\bf X}\right) + \iota_{\Sigma_K} \left({\bf X}\right) + \iota_{T_{\epsilon}} \left({\bf X}\right)\\
& \mbox{subject to} \; \left\|{\bf X} \right\|_2 = 1,
\end{split}
\end{equation}
where $T_{\epsilon}= \left\{{\bf X}\in \mathbb{R}^{N \times L} :\; \mbox{TV} \left({\bf X}\right)  \leq \epsilon \right\}$, with 
$\mbox{TV} \left({\bf x}\right) $ denoting the total variation (TV), which in the two-dimensional (2D) case is defined as
\begin{equation} \label{conv_TV}
\mbox{TV}({\bf X}) = \sum_{i=1}^{N-1}\sum_{j=1}^{L-1} \left(|X_{(i+1,j)} - X_{(i,j)}| + |X_{(i,j+1)} -X_{(i,j)}|\right),
\end{equation}
and $\epsilon$ is a positive parameter. In the same vein as 2DBIHT, the proposed algorithm is as follows:
{\begin{algorithm}{2DFBCS-TV}{
\label{alg:2DFBCS_TV}}
Set $\tau >0, \epsilon>0, K$, and ${\bf X}_0$ \\
\qrepeat\\
     ${\bf V}_{k+1} = {\bf X}_k  - \tau \,{\partial} f\left({\bf Y}\odot \left({\bf A}{\bf X}_k\right)\right)$\\
		  ${\bf X}_{k+1} =   {\mathcal P}_{\Sigma_K} \bigl( {\mathcal P}_{\mathcal{T}_{\epsilon}} \bigl({\bf V}_{k+1}\bigr)\bigr) $\\
					 $ k \leftarrow k+ 1$
\quntil some stopping criterion is satisfied.\\
\qreturn ${\bf X}/\left\|{\bf X}\right\|_2$
\end{algorithm}}
where line 4 is to compute the projection onto $T_{\epsilon}$, which can be obtained 
by using the algorithm proposed by Fadili and Peyr\'e \cite{fadili2011total}. The versions of the
2DBFCS-TV algorithm with $\ell_1$ and $\ell_2$ objectives are referred to as 
2DBFCS-TV-$\ell_1$ and 2DBFCS-TV-$\ell_2$, respectively.

\subsection{2DFBCS with Modified Total Variation}
We propose the following problem formulation of 2DFBCS:
\begin{equation} \label{2DBFCS_model2}
\begin{split}
& \min_{{\bf X}\in \mathbb{R}^{N\times L}}
 f\bigl({\bf Y}\odot \left({\bf A}{\bf X}\right)\bigr) + \iota_{\mathcal{F}^K_{\varepsilon}} \left({\bf X}\right)\\
& \mbox{subject\; to } \left\|{\bf X} \right\|_2 = 1,
\end{split}
\end{equation}
where the set $\mathcal{F}^K_{\varepsilon}$ requires a more careful explanation. As usual, define
${\Sigma_K} = \left\{{\bf X}\in \mathbb{R}^{N \times L} : \left\|{\bf X}\right\|_0 \leq K \right\}$ (with
$\|{\bf X}\|_0$ denoting the number of non-zeros in ${\bf X}$) as the set of $K$-sparse $N\times L$ images.
Consider the undirected 4-nearest-neighbors graph on the sites of $N\times L$ images, {\it i.e.},
$\mathcal{G}=(\mathcal{N},\mathcal{E})$, where $\mathcal{N} = \{(i,j),\, i=1,...,N, \, j=1,...,L\}$ and
$[(i,j),(k,l)]\in \mathcal{E}$ $\Leftrightarrow$
$\bigl((i=k)\wedge(|j-l|=1)\bigr)\vee \bigl((|i-k|=1)\wedge (j=l)\bigr)$. Given some ${\bf V} \in \mathbb{R}^{N \times L}$,
let $\widetilde{\mathcal{G}}({\bf V})=(\widetilde{\mathcal{N}}({\bf V}),\widetilde{\mathcal{E}}({\bf V}))$ be the
subgraph of $\mathcal{G}$ obtained by removing  all the nodes corresponding to zero
elements of ${\bf V}$ (that is, $(i,j) \in \widetilde{\mathcal{N}}({\bf V}) \Leftrightarrow V_{(i,j)}\neq 0$), as well as
the corresponding edges. Naturally, $\widetilde{\mathcal{G}}({\bf V})$ may not be a connected graph; define
$\{\mathcal{G}_1({\bf V}) ,...,\mathcal{G}_{\mathcal{K}({\bf V})}({\bf V}) \}$ as the set of the
$\mathcal{K}({\bf V})$ connected subgraphs of
$\widetilde{\mathcal{G}}({\bf V})$, where $\mathcal{G}_k({\bf V}) = \bigl( \mathcal{N}_k ({\bf V}),\mathcal{E}_k({\bf V})\bigr)$.
Define the normalized TV of the sub-image of ${\bf V}$ corresponding to each of these connected
subgraphs as 
\begin{equation} \label{normalizedTV}
\overline{\mbox{TV}}({ {\bf V}_{\mathcal{G}_k({\bf V})}}) =
|\mathcal{E}_k({\bf V})|^{-1} \sum_{\left[(i,j), (k,l)\right] \in \mathcal{E}_k({\bf V})} |V_{(i,j)} - V_{(k,l)}|
\end{equation}
 (assuming $|\mathcal{E}_k({\bf V})|>0$) where ${ {\bf V}_{\mathcal{G}_k({\bf V})}}$  is the subgraph indexed by 
$\mathcal{G}_k({\bf V})$.
Finally, the set $\mathcal{F}^K_{\varepsilon} \subseteq \Sigma_K $ is defined as
\begin{equation} \label{F_K}
\mathcal{F}^K_{\varepsilon} = \bigl\{ {\bf X} \in  \Sigma_K:\overline{\mbox{TV}}\bigl( {\bf X}_{{\mathcal{G}}_k({\bf X})} \bigr) \leq \varepsilon,\,
k=1,...,\mathcal{K}({\bf X}) \bigr\}
\end{equation}
In short, $\mathcal{F}^K_{\varepsilon}$ is the set of
$K$-sparse images such that the normalized TV of each of its connected blocks of non-zeros
doesn't exceed $\varepsilon$. Notice that this is different from the intersection of a
TV ball with $\Sigma_K$, as considered in \cite{kyrillidis2012hard}.

In the same vein as the 2DBIHT, We propose the following BIHT-type algorithm to solve (\ref{2DBFCS_model1}):
\begin{algorithm}{2DFBCS-MTV}{
\label{alg:2DFBCS_MTV}}
Set $\tau >0, \varepsilon>0, K$, and ${\bf X}_0$ \\
\qrepeat\\
     ${\bf V}_{k+1} = {\bf X}_k  - \tau \,{\partial} f\left({\bf Y}\odot \left({\bf A}{\bf X}_k\right)\right)$\\
		  ${\bf X}_{k+1} =    {\mathcal P}_{\mathcal{F}^K_{\varepsilon}} \bigl({\bf V}_{k+1}\bigr) $\\
					 $ k \leftarrow k+ 1$
\quntil some stopping criterion is satisfied.\\
\qreturn ${\bf X}/\left\|{\bf X}\right\|_2$
\end{algorithm}
In this algorithm, line 3 is also a sub-gradient descent step, where
${\partial} f$ is defined as \eqref{subgradient}
while line 4 performs the projection onto $\mathcal{F}^K_{\varepsilon}$. Although $\mathcal{F}^K_{\varepsilon}$ 
is non-convex, here we can briefly show that ${\mathcal P}_{\mathcal{F}^K_{\varepsilon}}$ 
can be computed as the follows (the details of computing ${\mathcal P}_{\mathcal{F}^K_{\varepsilon}}$
 are shown in Appendix): first, project onto $\Sigma_K$, {\it i.e.},
${\bf U} = \mathcal{P}_{\Sigma_K} ({\bf V}) $; then, ${\bf X} = {\mathcal P}_{\mathcal{F}^K_{\varepsilon}} \bigl({\bf V}\bigr)$ 
is obtained by projecting every connected group of non-zeros in ${\bf U}$ onto
the $\varepsilon$-radius normalized TV ball $\mathcal{B}_{\varepsilon}^{k}$:
\begin{equation} \label{B_TVbar}
\mathcal{B}_{\varepsilon}^k = 
\bigl\{ {\bf X}_{{{\mathcal{G}}_k({\bf X})}} \in \mathbb{R}^{{\mathcal{G}}_k({\bf X})}:
\overline{\mbox{TV}}\bigl( {\bf X}_{{\mathcal{G}}_k({\bf X})} \bigr) \leq \varepsilon, \bigr\}
\end{equation}  
for $k=1,...,\mathcal{K}({\bf X})$, {\it i.e.},
${\bf X}_{{\mathcal{G}}_k({\bf U})} = \mathcal{P}_{\mathcal{B}_{\varepsilon}^k}
\bigl({\bf U}_{{\mathcal{G}}_k({\bf U})}\bigr)$, for $k=1,...,\mathcal{K}({\bf U})$,
and keeping the zeros of ${\bf U}$, {\it i.e.}, ${\bf X}_{\mathcal{G}-\widetilde{\mathcal{G}}({\bf U})} =
{\bf U}_{\mathcal{G}-\widetilde{\mathcal{G}}({\bf U})}$.
Finally, as in \cite{jacques2011robust}, projection onto the unit sphere (line 6) satisfies the
constraint in (\ref{2DBFCS_model1}). The versions of the
2DBFCS-MTV algorithm with $\ell_1$ and $\ell_2$ objectives are referred to as 
2DBFCS-MTV-$\ell_1$ and 2DBFCS-MTV-$\ell_2$, respectively.

Of course, the objective functions in \eqref{2DBFCS_model1} and \eqref{2DBFCS_model2} are not convex 
(since $\Sigma_K$ is not a convex set and the $\{{\bf X}\in \mathbb{R}^{N \times L}: \; \|{\bf X}\|_2=1\}$ is also not a convex set),
 thus there is no guarantee that the algorithm finds a global 
minimum. If the original signal is known  to be non-negative, 
then the algorithm should include a projection onto $\mathbb{R}_+^{N \times L}$ in each iteration.

\section{Experiments}
In this section, we report results of experiments aimed at comparing the performance of 2DBFCS with that of 2DBIHT. 
Without loss of generality, we assume the original group-sparse image 
${\bf X} \in \mathbb{R}^{400 \times 100}$ , in which, 10 
line-groups are randomly generated, and each line-group has
9 elements valued by 10 or -10, and Then it is followed by a normalized operation ${\bf X} = {\bf X}/\left\|{\bf X}\right\|_2$. 
The sensing matrix ${\bf A}$ is a $200 \times 400$ matrix whose components are sampled from the standard normal distribution. 
And the variance of white Gaussian noise ${\bf W} \in \mathbb{R}^{200 \times 100}$ is $0.01$. Then the observations ${\bf Y}$ are obtained by (\ref{1bitcsnoisy}). 

\begin{figure}
	\centering
		\includegraphics[width=1\columnwidth]{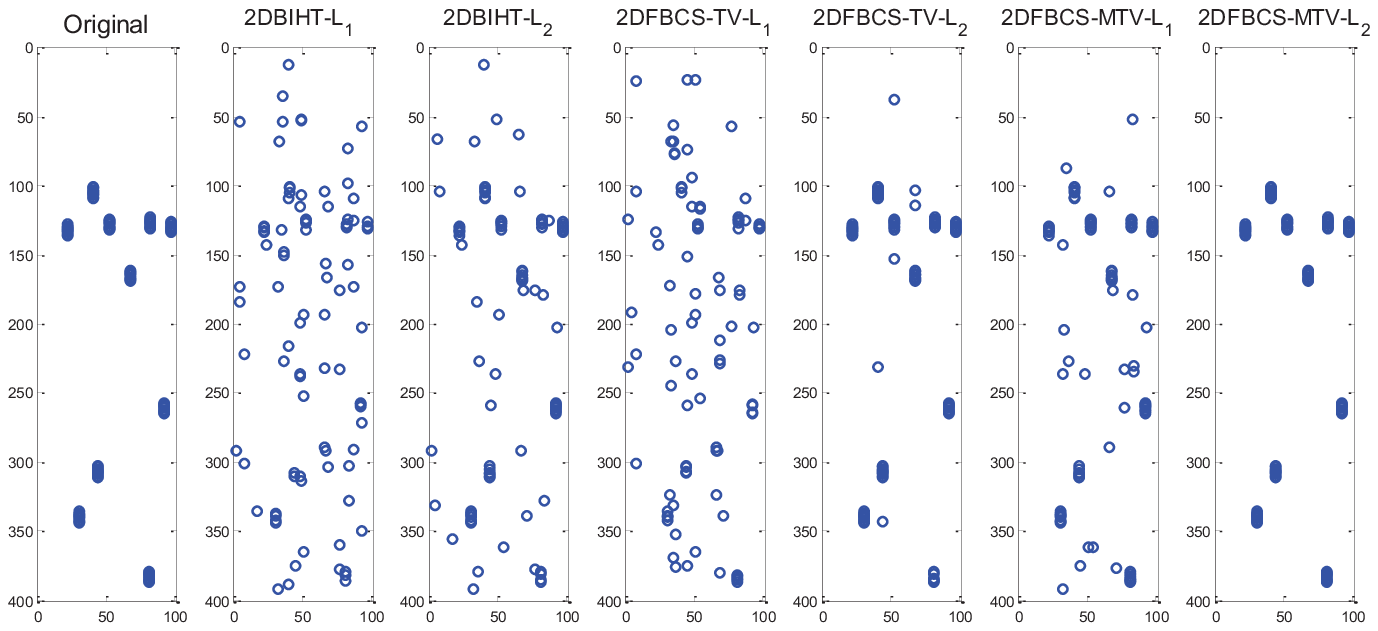}
	\caption{Recovered images by different algorithms}
	\label{fig:2DFBCS_10groups}
\end{figure}

We run the aforementioned six algorithms, the stepsizes of 2DBIHT-$\ell_1$ and 2DBIHT-$\ell_2$ are set as $\tau =1$ and $1/M$, respectively, and the 
parameters of  2DFBCS-TV-$\ell_1$, 2DFBCS-TV-$\ell_2$, 2DFBCS-MTV-$\ell_1$ and 2DFBCS-MTV-$\ell_2$ are hand tuned for the best improvement in signal-to-noise. 
The recovered signals are shown in Figure \ref{fig:2DFBCS_10groups}, from which, 
we can clearly see that the proposed 2DBFCS basically performs better than 2DBIHT. In general, the algorithms with the $\ell_2$ barrier outperforms
that with the $\ell_1$ barrier. Especially, the 2DFBCS-MTV-$\ell_2$ shows its superiority
over other algorithms, and nevertheless, the 2DFBCS-TV-$\ell_2$ is also good at recovering 
sparse piece-wise images. 

\section{Conclusions}

We have proposed the {\it 2D binary fused compressive sensing} (2DBFCS) to recover 2D sparse piece-wise smooth signals 
from 2D 1-bit compressive measurements. We have shown that if the original signals 
are in fact sparse and piece-wise smooth, the proposed method, 
is able to take advantage of the piece-wise smoothness of the original signal,  
 outperforms (under several accuracy measures) the 2D version of the previous method 
{\it binary iterative hard thresholding} (termed 2DBIHT), 
which  relies only on sparsity of the original signal. 
Future work will involve using the
 technique of detecting sign flips to obtain a robust version of 2DBFCS.

\bibliography{bibfile}
\end{document}